%% file: paper.tex
\pdfoutput=1
\documentclass[11pt]{article}

\usepackage[]{acl}
\usepackage{times}
\usepackage{latexsym}
\usepackage[T1]{fontenc}
\usepackage[utf8]{inputenc}
\usepackage{microtype}
\usepackage{graphicx}
\usepackage{booktabs}
\usepackage{multirow}
\usepackage{enumitem}
\usepackage{amsfonts}
\usepackage{amsmath}

\usepackage{bm}

\title{Debiasing Masks:\\A New Framework for Shortcut Mitigation in NLU}

\author{
Johannes Mario Meissner$^\dagger$, Saku Sugawara$^\ddagger$, Akiko Aizawa$^{\dagger\ddagger}$ \\
$^\dagger$The University of Tokyo, $^\ddagger$National Institute of Informatics \\
\texttt{\{meissner,saku,aizawa\}@nii.ac.jp}
}

\begin{document}
\maketitle
\begin{abstract}

Debiasing language models from unwanted behaviors in Natural Language Understanding tasks is a topic with rapidly increasing interest in the NLP community. Spurious statistical correlations in the data allow models to perform shortcuts and avoid uncovering more advanced and desirable linguistic features.
A multitude of effective debiasing approaches has been proposed, but flexibility remains a major issue. For the most part, models must be retrained to find a new set of weights with debiased behavior.
We propose a new debiasing method in which we identify debiased pruning masks that can be applied to a finetuned model. This enables the selective and conditional application of debiasing behaviors.
We assume that bias is caused by a certain subset of weights in the network; our method is, in essence, a mask search to identify and remove biased weights.
Our masks show equivalent or superior performance to the standard counterparts, while offering important benefits.
Pruning masks can be stored with high efficiency in memory, and it becomes possible to switch among several debiasing behaviors (or revert back to the original biased model) at inference time. Finally, it opens the doors to further research on how biases are acquired by studying the generated masks. For example, we observed that the early layers and attention heads were pruned more aggressively, possibly hinting towards the location in which biases may be encoded.

\end{abstract}

\section{Introduction}

The issue of spurious correlations in natural language understanding datasets has been extensively studied in recent years \cite{gururangan-etal-2018-annotation, mccoy-etal-2019-right, gardner-etal-2021-competency}. In the MNLI dataset \cite{williams-etal-2018-broad}, negation words such as ``not'' are unintended hints for the \textit{contradiction} label \cite{gururangan-etal-2018-annotation}, while a high word overlap between the premise and the hypothesis often correlates with \textit{entailment} \cite{mccoy-etal-2019-right}.

\begin{figure}[ht]
    \centering
    \includegraphics[scale=0.72]{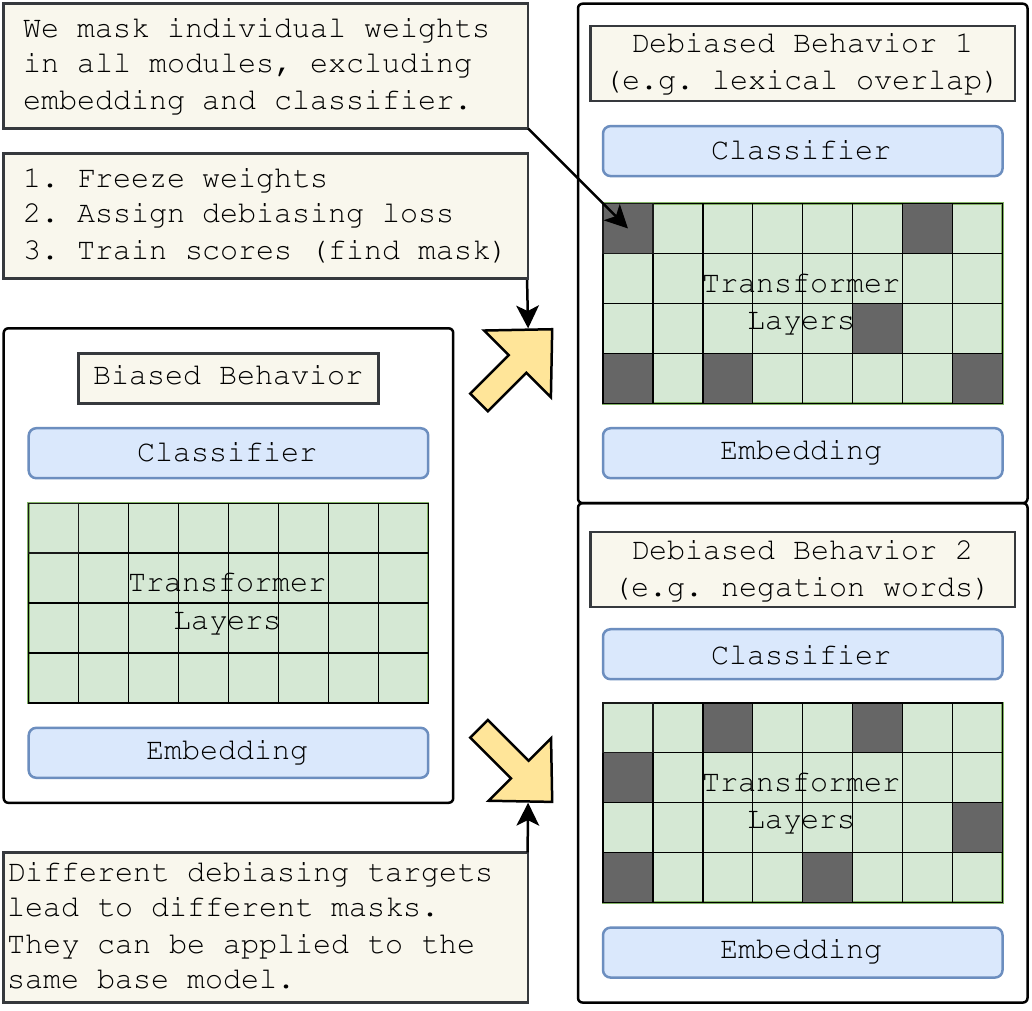}
    \caption{We find masks that remove bias from a finetuned model by using frozen-weight movement pruning.}
    \label{fig:approach}
\end{figure}

A common way to prevent the acquisition of biases (in this paper defined as unintended shortcut behaviors) is to adopt a debiasing loss function, such as product-of-experts \cite{hinton2002training}, to discourage learning shortcuts through the use of the annotated level of bias in each sample. Because manual bias annotation is difficult and expensive, the predictions of a biased model are used as bias annotations \cite{clark-etal-2019-dont}. Commonly, models must be retrained to achieve debiased behavior; this limits our ability to target different biases separately, as well as choose varied levels of debiasing strength (which impacts the in- vs. out-of-distribution trade-off).

We propose a new debiasing framework that focuses on removing bias from an existing model, instead of the now-common approach of re-training from scratch. Our approach is depicted in Figure~\ref{fig:approach}. We find a pruning mask that zeroes out the weights that cause biased outputs, producing the desired effect without altering the original model. This approach offers several clear advantages. First, pruning masks can be stored very efficiently, only occupying a fraction of the original model size. This enables creating multiple masks for varied debiasing behaviors. Secondly, masks can be effortlessly set or unset at inference time, as opposed to replacing the entire model. This allows for flexible application, as well as easy reversion to the original model when needed. Finally, re-framing the debiasing process as a mask search opens the doors towards future analysis directions, helping to more deeply understand how biases are learned, and how they can be eliminated.

\section{Related Work}

\subsection{Shortcuts in NLU}

The study of shortcut solutions in machine learning \cite{geirhos2020shortcut, d2020underspecification} has gained attention in recent years, including in the field of Natural Language Understanding. In SQuAD \citep{rajpurkar-etal-2016-squad}, \citet{jia-liang-2017-adversarial} show that distractor sentences can be inserted in such a way that the frequently used spurious features mislead the model's answer. In MNLI, \citet{gardner-etal-2021-competency} discuss the idea that all simple feature-label correlations should be regarded as spurious. Their results go in line with \citet{gururangan-etal-2018-annotation}, who show that models are able to achieve strong performance just by training on the hypothesis, a scenario where the desirable advanced features are not available. Instead, simple word correlations are used to make the prediction. Finally, other kinds of features such as lexical overlap have been pointed out as important shortcuts \citep{mccoy-etal-2019-right}.

\subsection{Debiasing} \label{sec:related-debiasing}

A wide range of debiasing approaches have been proposed to alleviate shortcut behavior.
For example, perturbations in the model's embedding space can encourage robustness to shortcuts \cite{liu2020adversarial}, and training on adversarial data \cite{wang-bansal-2018-robust} has important benefits for generalization capabilities too.
Removing a certain subset in the training data (filtering) can help to avoid learning spurious correlations. \citet{bras2020adversarial} devise an algorithm that selects samples to filter out, reducing the training time and robustness of the target model.

We will focus our efforts on a family of approaches that rely on a debiasing loss function to discourage biased learning, instead of altering the original training data. A wide range of debiasing losses have been proposed to discourage shortcut learning. They all rely on having a measure of the level of bias for each training sample, which is commonly obtained via a biased model's predictions. Among the most common are product-of-experts and focal loss \cite{schuster-etal-2019-towards}. Other approaches introduce a higher level of complexity, such as confidence regularization \cite{utama-etal-2020-mind} requiring a teacher model, or learned-mixin \cite{clark-etal-2019-dont} introducing an additional noise parameter.

\subsection{Pruning}

Pruning consists in masking out or completely removing weights in a neural network, such that they no longer contribute to the output. Common goals include reducing the model size or achieving an inference speed-up.

Magnitude pruning is a basic pruning method that removes weights based on their absolute value. It has proven to be an effective method to reduce model size without compromising on performance. \citet{gordon-etal-2020-compressing} apply this technique on transformer language models. Movement pruning, on the other hand, was proposed by \citet{sanh-2020-movement-pruning}, and involves training a score value alongside each weight. Scores are updated as part of the optimization process, and weights with an associated score below the threshold are masked out.

\citet{zhao-etal-2020-masking} introduce an alternative to the usual finetuning process by finding a mask for the pretrained base model such that performance on the target task increases. The same base model can be used with several masks to perform multiple tasks. They show that this approach achieves similar performance to standard finetuning on the GLUE tasks \cite{wang-etal-2018-glue}.

Our work is inspired by \citet{zhao-etal-2020-masking}; but we focus on removing biases from an already finetuned model. We will refer to this method as a mask search. It benefits from the same advantages: masks can be easily applied and removed from a shared base model, while additionally offering storage improvements.

\section{Masked Debiasing}

Our proposed approach to debiasing takes a unique point of view in the debiasing field by assuming that biased behavior is encoded in specific weights of the network and can be removed without altering the remaining weights. Thus, we perform a mask search to identify and remove those weights, experiencing debiased behavior in the resulting model. Our approach comes together by combining a debiasing loss, a biased model, and a score-based pruning technique. 

\paragraph{Debiasing Loss}

Among the debiasing losses mentioned in Section~\ref{sec:related-debiasing}, none appear to be clearly superior, each offering certain strengths and weaknesses, and striking a balance between in- and out-of-distribution performance. We err on the side of simplicity and run our experiments with the focal loss.
We follow \citet{clark-etal-2019-dont} for the implementation.

\begin{figure}[t]
    \centering
    \includegraphics[scale=0.66]{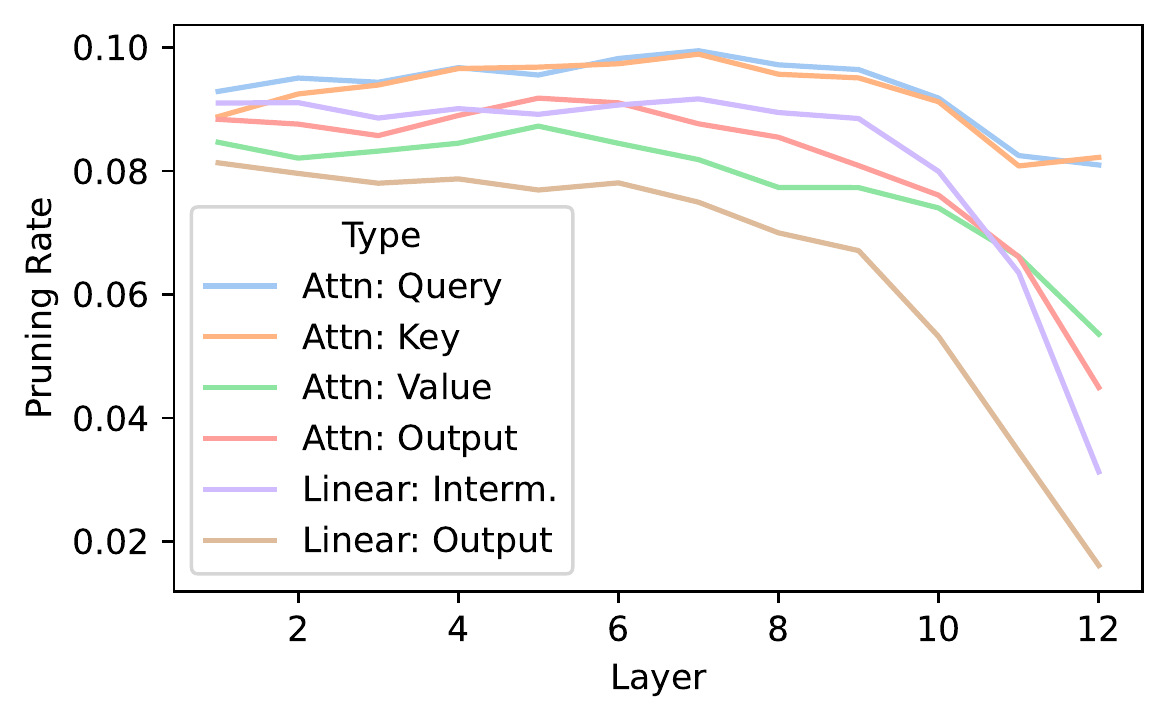}
    \caption{Pruning rate for each module in one of our debiasing masks.}
    \label{fig:pruning-rates}
\end{figure}

\paragraph{Bias Model}

To utilize a debiasing loss, we must obtain the predictions of a biased model on the training samples. We utilize weak learners, following the nomenclature introduced by \citet{sanh2021learning}; they do not require making any assumptions on the underlying biases. While our method works with any debiasing loss and bias level source, we choose this setting due to its flexibility and adaptability to new scenarios.

To date, two major learners have been proposed: undertrained \cite{utama-etal-2020-towards} and underparameterized \cite{sanh2021learning} learners. We train and extract the predictions of both of them for comparison. Undertrained learners are trained by selecting a very small subset of the training data, but keeping the target model's architecture and hyperparameters. This effectively translates into a model that overfits on the selected data subset. Features uncovered by the model in this manner are deemed spurious, as it is assumed that more advanced features require exploring larger quantities of data. We used 2000 samples for MNLI, and 500 samples for the other datasets. Underparameterized learners, on the other hand, restrict the model complexity by reducing layer size and count. The expectation is that this underparameterization is restrictive and limits the ability to find complex features. We use BERT-Tiny \cite{turc2019well}, a BERT model with 2 layers and an inner size of 128.

\paragraph{Frozen Movement Pruning}

Our approach is similar to \citet{zhao-etal-2020-masking}; but we implement it by leveraging the framework provided by \citet{sanh-2020-movement-pruning}, with the addition of weight-freezing. We use unstructured pruning, which means that each individual weight is considered for exclusion. Further, we utilize a threshold approach: in each weight tensor we remove (mask out) those weights with an associated score that is lower than the threshold. A regularization term for the scores helps avoid all scores growing larger than the threshold. For further details, we refer the reader to \citet{sanh-2020-movement-pruning}. As the starting point, we load a model already finetuned on the target task.

\section{Experimental Setup}

We compare our approach against the standard approach of re-training the model with a debiasing loss. Our experiments are carried out with BERT-Base \cite{devlin-etal-2019-bert}.

\input{table}

\subsection{Tasks and Evaluation Scenarios}

We perform experiments on three tasks in NLU, and consider several biases and evaluation datasets.

\paragraph{Natural Language Inference (NLI)} 
This task consists in classifying whether the relationship between a pair of sentences (premise and hypothesis) is entailed, neutral or contradicted. We evaluate on two well-known biases: lexical overlap bias \cite{mccoy-etal-2019-right} and negation words bias \cite{gururangan-etal-2018-annotation}. We train on MNLI, and evaluate on HANS \cite{mccoy-etal-2019-right} and our own negation-words subset. HANS can be used to evaluate the lexical overlap bias. Additionally, we create our own negation-words anti-bias set by selecting \textit{entailed} samples from the MNLI validation set that contain negation words in the hypothesis.\footnote{We select entailed samples with at least one of the following words in the hypothesis: no, not, don't, none, nothing, never, aren't, isn't, weren't, neither, don't, didn't, doesn't, cannot, hasn't won't.}

\paragraph{Paraphrase Identification}
In this task the goal is to identify whether a pair of sentences are paraphrasing each other or not. We train on QQP (Quora Question Pairs)\footnote{\url{https://www.kaggle.com/c/quora-question-pairs}}, and evaluate on PAWS \cite{zhang-etal-2019-paws}, which tests the model's reliability on the lexical overlap bias.

\paragraph{Fact Verification} We train on FEVER \cite{thorne-etal-2018-fever}, and evaluate on FEVER Symmetric \cite{schuster-etal-2019-towards}. The task setup is similar to NLI; we classify a claim-evidence pair as either support, refutes or not-enough-information. FEVER Symmetric eliminates claim-only biases (clues in the claim that allow to guess the label), among which are negation words, in a similar fashion to MNLI.

\subsection{Hyperparameters and Reproducibility}

We aim for complete reproducibility by providing complete code and clear reproduction instructions in our repository.\footnote{\url{https://github.com/mariomeissner/shortcut-pruning}} In most cases, we follow the configurations indicated by the respective original papers. Appendix~\ref{appendix:hyperparameters} is additionally provided for a report of important hyperparameters.

We run all of our experiments on five seeds. We not only seed the debiasing masks, but also the accompanying weak models and base models too.

\section{Results}

We compare our approach against our reproduction of the two weak model approaches, and compile our results in Table~\ref{tab:main-results}.

\subsection{Masked Debiasing is Effective}

First, we observe that our debiasing masks are very effective, surpassing the performance of their standard debiasing counterparts in two out of the three evaluated tasks, while keeping a competitive in-distribution performance. In the FEVER task, our masks provided a slight debiasing effect, but did not beat the standard debiasing baseline.
We do not report undertrained debiasing results in the QQP setting due to their failure to converge to adequate results (in both standard and masked debiasing).

To confirm that the combination of masking with a debiasing loss is necessary to achieve our results, we provide an ablation experiment (Masking w/o Debiasing) where the debiasing loss is replaced with the standard cross-entropy loss, while using the same pruning parameters. Results suggest that pruning alone is not able to achieve the same debiasing effect.

\subsection{Masks are Diverse}

As a means to showcase the capacity of debiasing masks to offer different debiasing behaviors, we refer to each method's capacity to mitigate the negation bias, evaluated by our negation-words subset.

As the standard debiasing results reveal, both weak model approaches were unable to improve baseline performance on our subset. Therefore, we used the hypothesis-only bias model, as provided by \citet{utama-etal-2020-mind}. Negation words are strongly correlated with the hypothesis-only bias; accordingly, observe a performance improvement on our subset when leveraging this debiasing target.

\section{Mask Analysis}

Our pruning approach uses a score threshold, which allows for a dynamic pruning rate across modules. To gain a better understanding on our masks, we study the generated density distribution. Specifically, we study the rate at which weights were removed in different network regions. In each layer, we obtain the pruning rate of the attention layer's query, key, value, and output modules, as well as the two linear modules that follow. We plot the results in Figure~\ref{fig:pruning-rates}.

We make two general observations. First, pruning rates are relatively higher in the early layers. It is known that the last few layers are very task specific \cite{tenney-etal-2019-bert}, which likely implies that their modification more directly impacts performance. Thus, our masks may be targeting early layers as a means to remove biases without causing a noticeable performance decrease. Secondly, the attention modules are more heavily pruned than the linear layers, which could suggest that attention heads play an important role in bias encoding. 

\section{Conclusion}

We conclude that debiasing masks are an effective approach to mitigating shortcuts. Alongside providing surprisingly well-performing debiased behaviors, masks allow to shift the way we think about debiasing: no longer should biased and unbiased models be treated as two separate models; rather, it becomes possible to ``remove'' biases by simply eliminating certain weights from the network.

\section*{Limitations}

An important limitation of this approach is that we found it necessary to follow \citet{sanh-2020-movement-pruning} and run the movement pruning technique for 12 epochs, requiring longer training time. We hypothesize that in future work, it could become possible to drastically reduce training time for this approach.

Further, in this short paper we explored three tasks in NLU with a fixed pruning configuration, but it would be beneficial to explore its applicability to other domains, combine it with other debiasing approaches, or explore more varied configurations, such as structured pruning methods for improved inference efficiency.

\section*{Acknowledgments}

The education period leading to this project has received funding from ``la Caixa'' Foundation (ID 100010434), under agreement LCF/BQ/AA19/11720042. This work was also supported by JST PRESTO Grant Number JPMJPR20C and JSPS KAKENHI Grant Number 21H03502.

\bibliography{anthology,custom}
\bibliographystyle{acl_natbib}

\appendix

\section{Negation Words} \label{appendix:negation-words}
Within the MNLI Validation Matched set, we select those entailed samples where at least one of the following words is present in the hypothesis: no, not, don't, none, nothing, never, aren't, isn't, weren't, neither, don't, didn't, doesn't, cannot, hasn't won't.

\section{Hyperparameters} \label{appendix:hyperparameters}

\subsection{Baseline and Debiased Models}

Our baseline models are trained with a batch size of 32, learning rate of 3e-5, weight decay of 0.1, and 20\% warmup steps.

\subsection{Weak Learners}

Undertrained models are trained with the standard BERT architecture and 2000 samples (500 in the case of QQP and FEVER) of the dataset for 5 epochs. \citet{utama-etal-2021-avoiding} mention 3 in their paper, but train with 5 in their code repository. We confirm that 5 yields better results. For underparameterized models, we follow \cite{sanh2021learning} and use BERT-Tiny \cite{turc2019well}, a BERT model with 2 layers and an inner size of 128. This model is trained on the full training set for 3 epochs, and slightly adjusted hyperparameters.

We use focal loss (sample reweighting) in all mask debiasing experiments, with the exception of the HypOnly MD model, which uses product-of-experts (in an attempt to follow \citet{utama-etal-2020-mind} as closely as possible).

\subsection{Masked Debiasing Models}

Our mask search is performed with a score learning rate of 0.1, batch size of 128, and 12 epochs of training.

\end{document}

%% file: table.tex
\begin{table*}[ht]
\centering
\resizebox{\textwidth}{!}{%
\begin{tabular}{@{}lcccccccc@{}}
\toprule
 &
  \multicolumn{3}{c}{MNLI} &
  \multicolumn{3}{c}{QQP} &
  \multicolumn{2}{c}{FEVER} \\
  \cmidrule(lr){2-4}
  \cmidrule(lr){5-7}
  \cmidrule(lr){8-9}
&
  \multicolumn{1}{c}{Val.} &
  \multicolumn{1}{c}{HANS} &
  \multicolumn{1}{c}{Neg.} &
  \multicolumn{1}{l}{Val.} &
  \multicolumn{1}{l}{PAWS} &
  \multicolumn{1}{l}{PAWS ($\neg$)} &
  \multicolumn{1}{l}{Val.} &
  \multicolumn{1}{l}{Symm.} \\ \midrule
Baseline &
  \textbf{84.38} &
  61.97 &
  78.47 &
  \textbf{92.07} &
  44.28 &
  23.37 &
  85.67 &
  64.08 \\
Masking w/o Debiasing &
  83.15 &
  64.16 &
  77.87 &
  90.91 &
  37.61 &
  16.38 &
  84.71 &
  63.26 \\
Undertrained SD &
  81.02 &
  67.06 &
  74.82 &
  \multicolumn{1}{c}{--} &
  \multicolumn{1}{c}{--} &
  \multicolumn{1}{c}{--} &
  \textbf{86.54} &
  65.06 \\
Underparam. SD &
  83.03 &
  67.75 &
  72.36 &
  88.79 &
  43.50 &
  27.16 &
  85.70 &
  \textbf{65.79} \\
Undertrained MD (Ours) &
  81.85 &
  \textbf{68.69} &
  75.58 &
  \multicolumn{1}{c}{--} &
  \multicolumn{1}{c}{--} &
  \multicolumn{1}{c}{--} &
  84.55 &
  64.85 \\
Underparam. MD (Ours) &
  82.24 &
  67.86 &
  73.95 &
  89.61 &
  \textbf{44.34} &
  \textbf{28.56} &
  84.96 &
  63.37 \\
HypOnly MD (Ours) &
  82.87 &
  64.26 &
  \textbf{79.03} &
  \multicolumn{1}{c}{--} &
  \multicolumn{1}{c}{--} &
  \multicolumn{1}{c}{--} &
  \multicolumn{1}{c}{--} &
  \multicolumn{1}{c}{--} \\
\bottomrule
\end{tabular}
}
\caption{Our experimental results comparing a BERT-Base baseline against several debiasing methods, along with our own mask debiasing approach. SD stands for standard debiasing, while MD stands for masked debiasing. MNLI Neg. is our split of the MNLI development set containing at least one negation word in the hypothesis. PAWS ($\neg$) indicates the subset of PAWS with a negative label (the anti-bias set for lexical overlap in QQP).}
\label{tab:main-results}
\end{table*}

%% file: paper.bbl
\begin{thebibliography}{27}
\expandafter\ifx\csname natexlab\endcsname\relax\def\natexlab#1{#1}\fi

\bibitem[{Bras et~al.(2020)Bras, Swayamdipta, Bhagavatula, Zellers, Peters,
  Sabharwal, and Choi}]{bras2020adversarial}
Ronan~Le Bras, Swabha Swayamdipta, Chandra Bhagavatula, Rowan Zellers, Matthew
  Peters, Ashish Sabharwal, and Yejin Choi. 2020.
\newblock \href {https://openreview.net/forum?id=H1g8p1BYvS} {Adversarial
  filters of dataset biases}.

\bibitem[{Clark et~al.(2019)Clark, Yatskar, and
  Zettlemoyer}]{clark-etal-2019-dont}
Christopher Clark, Mark Yatskar, and Luke Zettlemoyer. 2019.
\newblock \href {https://doi.org/10.18653/v1/D19-1418} {Don{'}t take the easy
  way out: Ensemble based methods for avoiding known dataset biases}.
\newblock In \emph{Proceedings of the 2019 Conference on Empirical Methods in
  Natural Language Processing and the 9th International Joint Conference on
  Natural Language Processing (EMNLP-IJCNLP)}, pages 4069--4082, Hong Kong,
  China. Association for Computational Linguistics.

\bibitem[{Devlin et~al.(2019)Devlin, Chang, Lee, and
  Toutanova}]{devlin-etal-2019-bert}
Jacob Devlin, Ming-Wei Chang, Kenton Lee, and Kristina Toutanova. 2019.
\newblock \href {https://doi.org/10.18653/v1/N19-1423} {{BERT}: Pre-training of
  deep bidirectional transformers for language understanding}.
\newblock In \emph{Proceedings of the 2019 Conference of the North {A}merican
  Chapter of the Association for Computational Linguistics: Human Language
  Technologies, Volume 1 (Long and Short Papers)}, pages 4171--4186,
  Minneapolis, Minnesota. Association for Computational Linguistics.

\bibitem[{D’Amour et~al.(2020)D’Amour, Heller, Moldovan, Adlam, Alipanahi,
  Beutel, Chen, Deaton, Eisenstein, Hoffman et~al.}]{d2020underspecification}
Alexander D’Amour, Katherine Heller, Dan Moldovan, Ben Adlam, Babak
  Alipanahi, Alex Beutel, Christina Chen, Jonathan Deaton, Jacob Eisenstein,
  Matthew~D Hoffman, et~al. 2020.
\newblock Underspecification presents challenges for credibility in modern
  machine learning.
\newblock \emph{Journal of Machine Learning Research}.

\bibitem[{Gardner et~al.(2021)Gardner, Merrill, Dodge, Peters, Ross, Singh, and
  Smith}]{gardner-etal-2021-competency}
Matt Gardner, William Merrill, Jesse Dodge, Matthew Peters, Alexis Ross, Sameer
  Singh, and Noah~A. Smith. 2021.
\newblock \href {https://doi.org/10.18653/v1/2021.emnlp-main.135} {Competency
  problems: On finding and removing artifacts in language data}.
\newblock In \emph{Proceedings of the 2021 Conference on Empirical Methods in
  Natural Language Processing}, pages 1801--1813, Online and Punta Cana,
  Dominican Republic. Association for Computational Linguistics.

\bibitem[{Geirhos et~al.(2020)Geirhos, Jacobsen, Michaelis, Zemel, Brendel,
  Bethge, and Wichmann}]{geirhos2020shortcut}
Robert Geirhos, J{\"o}rn-Henrik Jacobsen, Claudio Michaelis, Richard Zemel,
  Wieland Brendel, Matthias Bethge, and Felix~A Wichmann. 2020.
\newblock Shortcut learning in deep neural networks.
\newblock \emph{Nature Machine Intelligence}, 2(11):665--673.

\bibitem[{Gordon et~al.(2020)Gordon, Duh, and
  Andrews}]{gordon-etal-2020-compressing}
Mitchell Gordon, Kevin Duh, and Nicholas Andrews. 2020.
\newblock \href {https://doi.org/10.18653/v1/2020.repl4nlp-1.18} {Compressing
  {BERT}: Studying the effects of weight pruning on transfer learning}.
\newblock In \emph{Proceedings of the 5th Workshop on Representation Learning
  for NLP}, pages 143--155, Online. Association for Computational Linguistics.

\bibitem[{Gururangan et~al.(2018)Gururangan, Swayamdipta, Levy, Schwartz,
  Bowman, and Smith}]{gururangan-etal-2018-annotation}
Suchin Gururangan, Swabha Swayamdipta, Omer Levy, Roy Schwartz, Samuel Bowman,
  and Noah~A. Smith. 2018.
\newblock \href {https://doi.org/10.18653/v1/N18-2017} {Annotation artifacts in
  natural language inference data}.
\newblock In \emph{Proceedings of the 2018 Conference of the North {A}merican
  Chapter of the Association for Computational Linguistics: Human Language
  Technologies, Volume 2 (Short Papers)}, pages 107--112, New Orleans,
  Louisiana. Association for Computational Linguistics.

\bibitem[{Hinton(2002)}]{hinton2002training}
Geoffrey~E Hinton. 2002.
\newblock Training products of experts by minimizing contrastive divergence.
\newblock \emph{Neural computation}, 14(8):1771--1800.

\bibitem[{Jia and Liang(2017)}]{jia-liang-2017-adversarial}
Robin Jia and Percy Liang. 2017.
\newblock \href {https://doi.org/10.18653/v1/D17-1215} {Adversarial examples
  for evaluating reading comprehension systems}.
\newblock In \emph{Proceedings of the 2017 Conference on Empirical Methods in
  Natural Language Processing}, pages 2021--2031, Copenhagen, Denmark.
  Association for Computational Linguistics.

\bibitem[{Liu et~al.(2020)Liu, Cheng, He, Chen, Wang, Poon, and
  Gao}]{liu2020adversarial}
Xiaodong Liu, Hao Cheng, Pengcheng He, Weizhu Chen, Yu~Wang, Hoifung Poon, and
  Jianfeng Gao. 2020.
\newblock \href {http://arxiv.org/abs/2004.08994} {Adversarial training for
  large neural language models}.
\newblock \emph{arXiv:2004.08994}.

\bibitem[{McCoy et~al.(2019)McCoy, Pavlick, and Linzen}]{mccoy-etal-2019-right}
Tom McCoy, Ellie Pavlick, and Tal Linzen. 2019.
\newblock \href {https://doi.org/10.18653/v1/P19-1334} {Right for the wrong
  reasons: Diagnosing syntactic heuristics in natural language inference}.
\newblock In \emph{Proceedings of the 57th Annual Meeting of the Association
  for Computational Linguistics}, pages 3428--3448, Florence, Italy.
  Association for Computational Linguistics.

\bibitem[{Rajpurkar et~al.(2016)Rajpurkar, Zhang, Lopyrev, and
  Liang}]{rajpurkar-etal-2016-squad}
Pranav Rajpurkar, Jian Zhang, Konstantin Lopyrev, and Percy Liang. 2016.
\newblock \href {https://doi.org/10.18653/v1/D16-1264} {{SQ}u{AD}: 100,000+
  questions for machine comprehension of text}.
\newblock In \emph{Proceedings of the 2016 Conference on Empirical Methods in
  Natural Language Processing}, pages 2383--2392, Austin, Texas. Association
  for Computational Linguistics.

\bibitem[{Sanh et~al.(2021)Sanh, Wolf, Belinkov, and Rush}]{sanh2021learning}
Victor Sanh, Thomas Wolf, Yonatan Belinkov, and Alexander~M Rush. 2021.
\newblock \href {https://openreview.net/forum?id=Hf3qXoiNkR} {Learning from
  others' mistakes: Avoiding dataset biases without modeling them}.
\newblock In \emph{International Conference on Learning Representations}.

\bibitem[{Sanh et~al.(2020)Sanh, Wolf, and Rush}]{sanh-2020-movement-pruning}
Victor Sanh, Thomas Wolf, and Alexander~M. Rush. 2020.
\newblock Movement pruning: Adaptive sparsity by fine-tuning.
\newblock NIPS'20, Red Hook, NY, USA. Curran Associates Inc.

\bibitem[{Schuster et~al.(2019)Schuster, Shah, Yeo, Roberto Filizzola~Ortiz,
  Santus, and Barzilay}]{schuster-etal-2019-towards}
Tal Schuster, Darsh Shah, Yun Jie~Serene Yeo, Daniel Roberto Filizzola~Ortiz,
  Enrico Santus, and Regina Barzilay. 2019.
\newblock \href {https://doi.org/10.18653/v1/D19-1341} {Towards debiasing fact
  verification models}.
\newblock In \emph{Proceedings of the 2019 Conference on Empirical Methods in
  Natural Language Processing and the 9th International Joint Conference on
  Natural Language Processing (EMNLP-IJCNLP)}, pages 3419--3425, Hong Kong,
  China. Association for Computational Linguistics.

\bibitem[{Tenney et~al.(2019)Tenney, Das, and Pavlick}]{tenney-etal-2019-bert}
Ian Tenney, Dipanjan Das, and Ellie Pavlick. 2019.
\newblock \href {https://doi.org/10.18653/v1/P19-1452} {{BERT} rediscovers the
  classical {NLP} pipeline}.
\newblock In \emph{Proceedings of the 57th Annual Meeting of the Association
  for Computational Linguistics}, pages 4593--4601, Florence, Italy.
  Association for Computational Linguistics.

\bibitem[{Thorne et~al.(2018)Thorne, Vlachos, Christodoulopoulos, and
  Mittal}]{thorne-etal-2018-fever}
James Thorne, Andreas Vlachos, Christos Christodoulopoulos, and Arpit Mittal.
  2018.
\newblock \href {https://doi.org/10.18653/v1/N18-1074} {{FEVER}: a large-scale
  dataset for fact extraction and {VER}ification}.
\newblock In \emph{Proceedings of the 2018 Conference of the North {A}merican
  Chapter of the Association for Computational Linguistics: Human Language
  Technologies, Volume 1 (Long Papers)}, pages 809--819, New Orleans,
  Louisiana. Association for Computational Linguistics.

\bibitem[{Turc et~al.(2019)Turc, Chang, Lee, and Toutanova}]{turc2019well}
Iulia Turc, Ming{-}Wei Chang, Kenton Lee, and Kristina Toutanova. 2019.
\newblock \href {http://arxiv.org/abs/1908.08962} {Well-read students learn
  better: The impact of student initialization on knowledge distillation}.
\newblock \emph{CoRR}, abs/1908.08962.

\bibitem[{Utama et~al.(2021)Utama, Moosavi, Sanh, and
  Gurevych}]{utama-etal-2021-avoiding}
Prasetya Utama, Nafise~Sadat Moosavi, Victor Sanh, and Iryna Gurevych. 2021.
\newblock \href {https://doi.org/10.18653/v1/2021.emnlp-main.713} {Avoiding
  inference heuristics in few-shot prompt-based finetuning}.
\newblock In \emph{Proceedings of the 2021 Conference on Empirical Methods in
  Natural Language Processing}, pages 9063--9074, Online and Punta Cana,
  Dominican Republic. Association for Computational Linguistics.

\bibitem[{Utama et~al.(2020{\natexlab{a}})Utama, Moosavi, and
  Gurevych}]{utama-etal-2020-mind}
Prasetya~Ajie Utama, Nafise~Sadat Moosavi, and Iryna Gurevych.
  2020{\natexlab{a}}.
\newblock \href {https://doi.org/10.18653/v1/2020.acl-main.770} {Mind the
  trade-off: Debiasing {NLU} models without degrading the in-distribution
  performance}.
\newblock In \emph{Proceedings of the 58th Annual Meeting of the Association
  for Computational Linguistics}, pages 8717--8729, Online. Association for
  Computational Linguistics.

\bibitem[{Utama et~al.(2020{\natexlab{b}})Utama, Moosavi, and
  Gurevych}]{utama-etal-2020-towards}
Prasetya~Ajie Utama, Nafise~Sadat Moosavi, and Iryna Gurevych.
  2020{\natexlab{b}}.
\newblock \href {https://doi.org/10.18653/v1/2020.emnlp-main.613} {Towards
  debiasing {NLU} models from unknown biases}.
\newblock In \emph{Proceedings of the 2020 Conference on Empirical Methods in
  Natural Language Processing (EMNLP)}, pages 7597--7610, Online. Association
  for Computational Linguistics.

\bibitem[{Wang et~al.(2018)Wang, Singh, Michael, Hill, Levy, and
  Bowman}]{wang-etal-2018-glue}
Alex Wang, Amanpreet Singh, Julian Michael, Felix Hill, Omer Levy, and Samuel
  Bowman. 2018.
\newblock \href {https://doi.org/10.18653/v1/W18-5446} {{GLUE}: A multi-task
  benchmark and analysis platform for natural language understanding}.
\newblock In \emph{Proceedings of the 2018 {EMNLP} Workshop {B}lackbox{NLP}:
  Analyzing and Interpreting Neural Networks for {NLP}}, pages 353--355,
  Brussels, Belgium. Association for Computational Linguistics.

\bibitem[{Wang and Bansal(2018)}]{wang-bansal-2018-robust}
Yicheng Wang and Mohit Bansal. 2018.
\newblock \href {https://doi.org/10.18653/v1/N18-2091} {Robust machine
  comprehension models via adversarial training}.
\newblock In \emph{Proceedings of the 2018 Conference of the North {A}merican
  Chapter of the Association for Computational Linguistics: Human Language
  Technologies, Volume 2 (Short Papers)}, pages 575--581, New Orleans,
  Louisiana. Association for Computational Linguistics.

\bibitem[{Williams et~al.(2018)Williams, Nangia, and
  Bowman}]{williams-etal-2018-broad}
Adina Williams, Nikita Nangia, and Samuel Bowman. 2018.
\newblock \href {https://doi.org/10.18653/v1/N18-1101} {A broad-coverage
  challenge corpus for sentence understanding through inference}.
\newblock In \emph{Proceedings of the 2018 Conference of the North {A}merican
  Chapter of the Association for Computational Linguistics: Human Language
  Technologies, Volume 1 (Long Papers)}, pages 1112--1122, New Orleans,
  Louisiana. Association for Computational Linguistics.

\bibitem[{Zhang et~al.(2019)Zhang, Baldridge, and He}]{zhang-etal-2019-paws}
Yuan Zhang, Jason Baldridge, and Luheng He. 2019.
\newblock \href {https://doi.org/10.18653/v1/N19-1131} {{PAWS}: Paraphrase
  adversaries from word scrambling}.
\newblock In \emph{Proceedings of the 2019 Conference of the North {A}merican
  Chapter of the Association for Computational Linguistics: Human Language
  Technologies, Volume 1 (Long and Short Papers)}, pages 1298--1308,
  Minneapolis, Minnesota. Association for Computational Linguistics.

\bibitem[{Zhao et~al.(2020)Zhao, Lin, Mi, Jaggi, and
  Sch{\"u}tze}]{zhao-etal-2020-masking}
Mengjie Zhao, Tao Lin, Fei Mi, Martin Jaggi, and Hinrich Sch{\"u}tze. 2020.
\newblock \href {https://doi.org/10.18653/v1/2020.emnlp-main.174} {Masking as
  an efficient alternative to finetuning for pretrained language models}.
\newblock In \emph{Proceedings of the 2020 Conference on Empirical Methods in
  Natural Language Processing (EMNLP)}, pages 2226--2241, Online. Association
  for Computational Linguistics.

\end{thebibliography}
